\documentclass[conference]{IEEEtran}

\usepackage{graphicx} 
\usepackage{mathptmx} 
\usepackage{times} 
\usepackage{amsmath} 
\usepackage{amssymb}  
\usepackage{hyperref}
\hypersetup{
	pdftitle={Building Generalization Into Behavior Generation Via Adaptive Compositions of Regularities},
	pdfsubject={Generalization in robotics through adaptive composition of regularities},
	pdfauthor={Aravind Battaje, Malte Bernhard, Vito Mengers, Oliver Brock},
	pdfkeywords={Generalization, Compositionality, Behavior generation, Active perception, Control and Dynamics, State Estimation, Modeling and Simulation}
}

\usepackage{microtype} 
\microtypesetup{nopatch=footnote}

\usepackage{siunitx}
\usepackage{setspace}
\usepackage{flushend}

\usepackage{cleveref}
\usepackage[noadjust,compress,sort,nospace,move]{cite}

\Crefname{figure}{Fig.}{Figs.}
\Crefname{section}{Sec.}{Secs.}
\Crefname{equation}{Eq.}{Eqs.}

\usepackage{color}
\definecolor{purple}{rgb}{0.5,0,0.5}
\definecolor{orange}{rgb}{1,0.5,0}
\definecolor{green}{rgb}{0,0.5,0}
\definecolor{blue}{rgb}{0,0,1}

\title{
\LARGE \bf Building Generalization Into Behavior Generation Via\\Adaptive Compositions of Regularities
}

\author{
\IEEEauthorblockN{Aravind Battaje\textsuperscript{1,2}\quad Malte Bernhard\textsuperscript{1}\quad Vito Mengers\textsuperscript{1,2}\quad Oliver Brock\textsuperscript{1,2,3}\\ \\}
\IEEEauthorblockA{\textsuperscript{1}Robotics and Biology Laboratory, Technische Universit\"at Berlin, Germany\\
\textsuperscript{2}Science of Intelligence, Research Cluster of Excellence, Berlin, Germany\\
\textsuperscript{3}Robotics Institute Germany}
}

\begin{document}

\maketitle

\begin{abstract}

Generalization in robotics requires prior knowledge about how the world is structured, yet this structure changes from one situation to the next. This paper investigates the proposition that generalization arises from \emph{adaptively composing regularities}---predictable relationships within the robot-environment system---into situation-appropriate structures for behavior generation. We examine this proposition by analyzing the mechanism in AICON (Active InterCONnect), a framework representing regularities as interacting processes in a differentiable network, where sensory feedback realizes composition and gradient descent generates behavior. To isolate adaptive composition as the key mechanism, we study a simple simulated problem in which all relevant regularities can be identified. We expose the resulting model to a wide range of novel conditions not considered during design, and we find that it generates context-appropriate behavior in all but one case, where encoded regularities are provably insufficient. Ablations reveal that the network automatically modulates which regularities influence behavior based on their informativeness. These results suggest that adaptive composition of regularities constitutes a powerful inductive bias for building generalization into behavior generation.

\end{abstract}


\section{Introduction}

Generalization is perhaps the central challenge in robotics. Every paradigm, from control and planning to learning, ultimately strives toward it. Solving generalization would, in a sense, mean solving robotics itself. But there is an obvious yet fundamental problem: a robot that generalizes must respond appropriately to a vast, possibly infinite space of situations. No designer could explicitly anticipate them all.

Yet this is not insurmountable. The world, despite its complexity, is not arbitrary. Its structure arises from separable factors~\cite{bengio_representation_2013,locatello2019challenging,higgins2017beta,liu_learning_2023,zhou_robodreamer_2024} that combine in different ways across situations. A robot equipped with representations of these factors can recombine them to handle situations never explicitly encountered. This is the essence of generalization through composition~\cite{lake_generalization_2018,brooks_robust_1986,ijspeert_dynamical_2013,wiedemer_compositional_2023}, which raises two questions: what are the relevant factors, and how should they be combined?

Many approaches attempt to discover such factors from data~\cite{kingma_auto-encoding_2014,sontakke_causal_2021}. In robotics, however, the relevant factors are typically not latent variables but physical relationships we already understand: inertia, kinematics, geometric constraints, and so on. We call these predictable relationships \emph{regularities}. Each regularity constrains how part of the world evolves. Together, they define a structured space of valid robot-environment configurations within which all feasible behaviors reside.

\begin{figure}[t]
	\centering
	\includegraphics[width=0.95\linewidth]{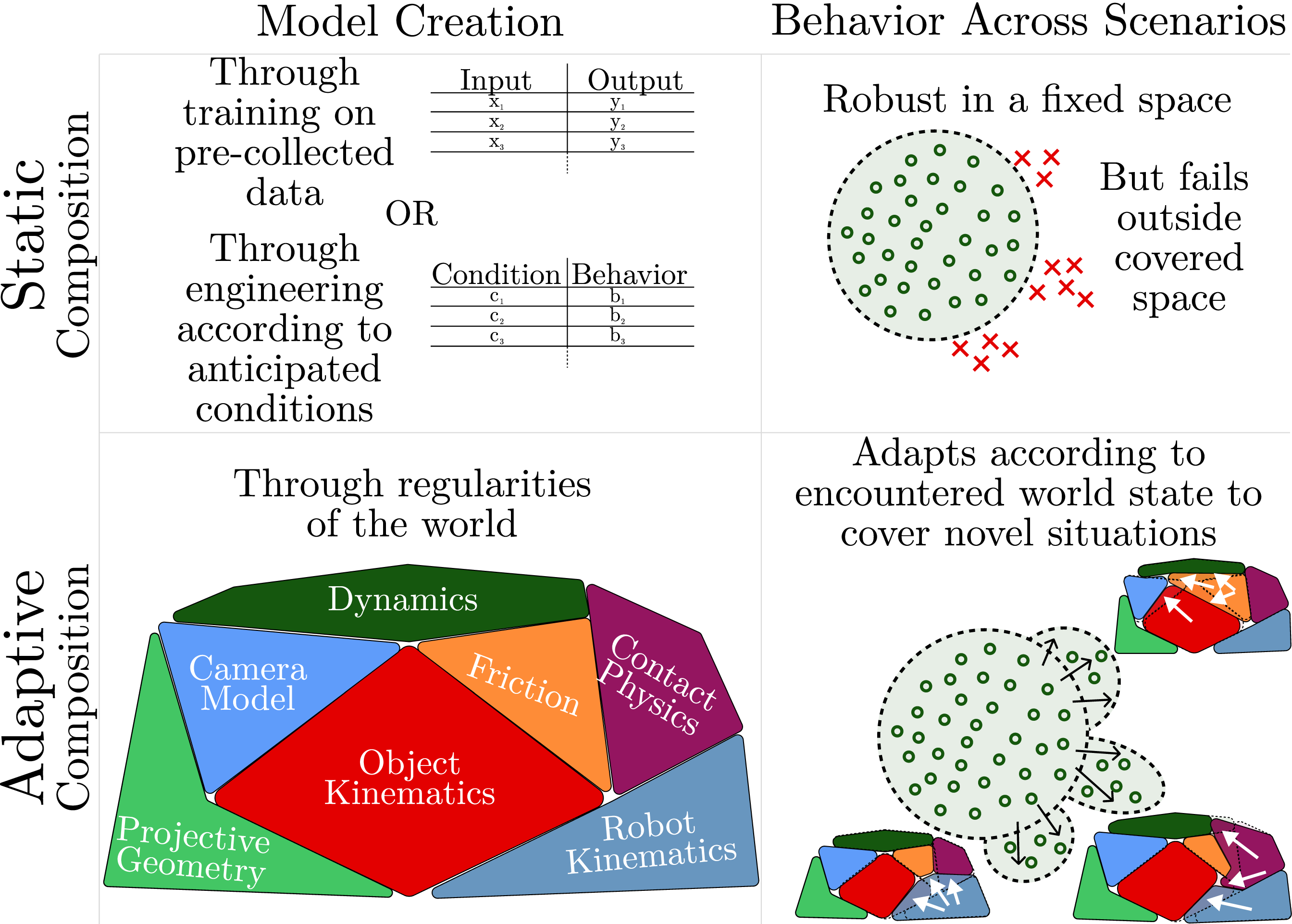}
	\caption{Adaptive composition enables generalization by reconfiguring regularities to match each situation. \textbf{(Top)} Whether learned from data or engineered for anticipated conditions, traditional approaches yield static compositions that cover only a bounded region---the training distribution or the set of designed-for conditions---and fail outside it (red crosses). \textbf{(Bottom)} Adaptive composition represents regularities (dynamics, kinematics, geometry, etc.) as interacting components that recombine based on sensory feedback. As the world state changes, the composition reconfigures, extending coverage to novel situations that no single fixed arrangement could handle.}
	\label{fig:title-fig}
\end{figure}

The challenge is not identifying regularities but combining them appropriately. But this problem contains its own solution: as the world changes, so does the relevant combination of regularities---and this change is observable through sensory feedback. 
A representation of regularities can therefore adapt its combination continuously. We call this an \emph{adaptive composition} of regularities (see \Cref{fig:title-fig}).

An adaptive composition is not a serial chain of functions, but a network of interacting influences that reflects how regularities couple in the physical world. As sensory feedback flows through this network, the influence of each regularity adapts to its momentary informativeness. Such an adaptive composition need only be specified once at the level of regularities, rather than redesigned for each case.

We examine adaptive composition using AICON (for \textbf{A}ctive \textbf{I}nter\textbf{con}nect)~\cite{battaje_information_2024,mengers_no_2025,mengers_leveraging_2024,mengers_robotics-inspired_2025}, a framework that encodes regularities as interacting processes in a differentiable network. Gradients through this network identify effective actions~\cite{mengers_no_2025}, enabling complex behavior without explicit search or switching. Earlier applications have demonstrated AICON's effectiveness across diverse robotics problems. Here, we examine its underlying mechanism and how it enables generalization.

To isolate adaptive composition as the key mechanism, we study a problem in which all relevant regularities can be identified. We then expose the resulting AICON model to conditions never considered during design, including moving targets, obstacles, sensorimotor disturbances, and modified sensing and actuation, as well as combinations of these. 

AICON generates context-appropriate behavior in 16 of 17 scenarios, automatically modulating which regularities dominate based on their informativeness. The single failure occurs when the encoded regularities are provably insufficient: acceleration control without velocity measurements leaves the integration constant unobservable. These results suggest that adaptive composition of regularities offers a practical path to building generalization into robot behavior.

\section{Related Work}\label{sec:related-work}
To situate adaptive composition as implemented in AICON within prior work, we invoke the concept of \emph{inductive bias}: the assumptions a model makes to enable generalization~\cite{mitchell_need_1980,mitchell_machine_1997}. Different behavior-generation paradigms embody different inductive biases, leading to distinct capabilities and limitations. We review these paradigms and contrast their assumptions with those of AICON.

\emph{Planning approaches}~\cite{fikes_strips_1971, lavalle2006planning,elbanhawi_sampling-based_2014,li_reactive_2021,zhang_symbolic_2023,pan_task_2024} assume that task-relevant structure can be captured through explicit state spaces and transition models, reducing behavior generation to path search. This bias works well in domains with stable, long-horizon structure---chess, logistics, geometric planning---but becomes brittle when robot-environment dynamics are context-dependent or long-term prediction is unreliable~\cite{mengers_no_2025,baum_world_2022}. These methods rely on a largely \emph{static composition}: the decomposition into states, actions, constraints, and dynamics is fixed at design time. Extensions such as belief-space~\cite{kaelbling_planning_1998,kaelbling_integrated_2013,platt_jr_belief_2010} or contingency planning~\cite{rhinehart_contingencies_2021} relax uncertainty assumptions, but still operate over a predefined set of model compositions. AICON performs continuous estimation through dense sensing and local updates. This allows the influence of regularities to adapt online and reduces reliance on prediction or hand-designed abstractions.

\emph{Model-based control} shares with AICON the assumption that observations map to actions through models of the robot-environment relationship. Traditional approaches encode this relationship through direct mappings~\cite{hutchinson_tutorial_1996}, monolithic state-based models~\cite{khatib_real-time_1986,zhao_hybrid_2004,yao_singularity-free_2021}, or model-predictive control over a unified forward model~\cite{mayne_model_2014,bhardwaj_storm_2021}. Even when multiple models or constraints are present, their interaction is typically fixed or manually scheduled, yielding a static composition. AICON instead encodes the robot-environment relationship as a factorized network of multiple regularities, deriving behavior from this model using gradient descent. This bears surface resemblance to one-step MPC, in that both select actions by gradient-based optimization of a one-step objective. But the substantive difference is structural. AICON's gradient aggregates contributions from multiple pathways rather than from a single forward model, so the relative influence of each regularity adapts online. Producing comparable context-sensitivity from a unified MPC model would require explicit switching logic or sufficiently long prediction horizons. Further, states in AICON are estimated recursively and history retained in beliefs, supporting robust integration across modalities and helping resolve perceptual aliasing.

\emph{Learning approaches} similarly assume observation-to-action mappings but often impose minimal inductive bias beyond generic architectural constraints. Across imitation~\cite{fishman_avoid_2025,chi_diffusion_2025}, reinforcement~\cite{schulman_proximal_2017,nagabandi_neural_2018,ibarz_how_2021}, and predictive model learning~\cite{deisenroth_pilco_2011,byravan_se3-pose-nets:_2018,groth_goal-conditioned_2021}, structure is largely inferred from data. Recent work emphasizes composition through disentangled representations~\cite{higgins2017beta}. But without strong inductive biases, learned decompositions often fail to support compositional generalization~\cite{locatello2019challenging,montero2022lost}. The result is interpolation within the training distribution~\cite{mitchell_need_1980} and failure under novel conditions~\cite{he_demystifying_2025}. AICON takes a different approach: by explicitly encoding regularities at an abstraction amenable to adaptive composition, it builds in an inductive bias that better supports robot behavior generation. The combination of regularities is then resolved online through interaction, which has yielded robustness across diverse applications~\cite{battaje_information_2024,mengers_no_2025,mengers_leveraging_2024,mengers_robotics-inspired_2025}.

\emph{Structural biases} cut across paradigms and shape generalization. Recurrent structures---recursive estimators~\cite{thrun_probablistic_2005}, smoothers~\cite{yi_differentiable_2021}, recurrent neural networks~\cite{feng_were_2024}---assume temporal smoothness. Relational structures---probabilistic graphical models~\cite{koller_probabilistic_2009}, factor graphs~\cite{dellaert_factor_2017}, Bayesian networks~\cite{yedidia_understanding_2003}, relational networks~\cite{cook_interacting_2011,martel_toward_2015}, graph neural networks~\cite{sanchez-gonzalez_learning_2020}---assume decomposable dependencies among components. These biases support compositional inference, but typically assume fixed interaction patterns. AICON combines recurrent and relational structure while allowing the effective composition of regularities to adapt continuously through feedback.

In summary, existing approaches either fix the composition of structure at design time or leave it to be discovered from data. AICON occupies a middle ground: it encodes regularities explicitly but allows their composition to adapt online.

\section{Adaptive Composition of Regularities}\label{sec:adaptive-composition}

In this section, we provide a geometric intuition for adaptive composition of regularities and their role in generalization. We begin by making our key terms precise.

A \emph{regularity} is a reproducible and predictable physical relationship between quantities in the robot-environment system. A system \emph{generalizes} if it produces task-appropriate behavior under conditions not considered during design or training, without redesign or retraining.

Regularities describe structure in the world. Inertia maintains the velocity of a rolling ball, gravity accelerates falling objects, and robot kinematics determine the position of its end-effector. Each regularity defines a manifold along which a part of the world evolves. For example, inertia constrains a rolling ball to a straight-line trajectory.

When a robot needs to act, it can utilize the structure these manifolds provide. Intercepting a rolling ball means finding a path that brings the end-effector to the same point in space-time as the ball's trajectory. This can be seen as linking two manifolds---ball dynamics and manipulator kinematics---and finding a suitable path through their intersection. Robustness follows naturally: minor disturbances shift the manifolds slightly, but valid paths persist.

Generalization, however, demands more than robustness. Different situations require different arrangements of manifolds. Intercepting a rolling ball, a falling ball, or a ball on an incline each requires a different combination of regularities. A system that generalizes must navigate through these dynamic arrangements. Moreover, the relevance of individual regularities shifts with context. If an obstacle partially blocks the interception path, collision-avoidance constraints suddenly dominate---but when the path is clear, this regularity contributes nothing. The composition must adapt, emphasizing whichever regularities are most informative at each moment. This is what we mean by \emph{adaptive composition}: not a fixed combination of regularities, but one that reconfigures as sensory feedback reveals which relationships currently govern the situation.

This perspective also clarifies a limit: a system can generalize only within the scope of its encoded regularities. Given such regularities, how does one implement an adaptive composition? AICON provides an answer as we describe next.

\section{AICON: A Framework for Adaptive Composition}\label{sec:method}

The previous section described adaptive composition geometrically: regularities define manifolds, and behavior emerges from navigating arrangements of manifolds that shift with context. AICON (Active Intercon\-nect)~\cite{battaje_information_2024,mengers_no_2025} implements this idea by representing regularities as interacting processes in a differentiable network. Sensory feedback realizes adaptive composition, while gradient descent generates behavior. Here, we describe AICON's core features (for details, see~\cite{mengers_no_2025}): first, two types of regularities (\Cref{sec:method-regularities}); then their expression in AICON's structural components (\Cref{sec:method-components}); and finally how behavior is generated through gradient descent (\Cref{sec:method-gradients}).

\subsection{Two Types of Regularities}\label{sec:method-regularities}
Section~\ref{sec:adaptive-composition} described regularities as manifolds. In AICON, these manifolds are expressed as constraints, with two types distinguished by what they constrain and implemented in distinct structural components.

\emph{Temporal regularities} describe how quantities evolve, e.g., how object positions change consistent with velocities. These constraints specify trajectories through time.

\emph{Cross-variable regularities} describe instantaneous relationships, e.g., how a rigid link dictates poses of successive joints. These constraints specify which combinations of variables are mutually consistent at each moment.

\subsection{Structural Components}\label{sec:method-components}

\subsubsection*{Recursive Estimators}
These encode temporal regularities. Each estimator maintains a belief over some state variable and updates it according to a differentiable dynamics model:
\begin{equation}
	f(\mathbf{x}_{t-1}; c_1, c_2, \ldots, c_N) = \mathbf{x}_t
\end{equation}
where \(\mathbf{x}_t\) is the state at time \(t\), \(f\) specifies how it evolves, and optional priors \(c_1, \ldots, c_N\) constrain the estimate. In our implementation, extended Kalman filters serve as recursive estimators, though other schemes (e.g., particle filters or categorical beliefs) are equally compatible~\cite{mengers_combining_2023}. Each estimator enforces temporal consistency, which computationally corresponds to constraining trajectories along a temporal manifold.

\subsubsection*{Active Interconnections}
These encode cross-variable regularities. Each interconnection expresses a lawful relationship among multiple state variables and sensorimotor signals:
\begin{equation}
	c(\mathbf{x}_1, \mathbf{x}_2, \ldots, \mathbf{x}_M) = \mathbf{0}
\end{equation}
where \(c\) captures a predictable dependency. Active interconnections propagate information bidirectionally: the constraint mediates information exchange across all connected variables.

Together, recursive estimators and active interconnections form a differentiable network that continuously adapts the effective combination of regularities based on sensory feedback. This network does more than estimate state: it embodies the compositional structure of the robot-environment relationship and makes it available for action selection.

\subsection{Generating Behavior Through Gradient Descent}\label{sec:method-gradients}

The differentiable structure of AICON exposes how actions propagate through the network to affect task-relevant quantities. This enables a direct approach to behavior generation: define a goal function \(g\) over state variables, and compute its gradient with respect to motor commands \(\mathbf{a}\):
\begin{equation}
	\mathbf{a}_{t+1} = \mathbf{a}_t - \mathbf{k} \cdot \nabla g(\mathbf{a}_t)
	\label{eqn:gradient-update}
\end{equation}
where \(\mathbf{k}\) is a gain parameter\footnote{See \Cref{sec:related-work} for comparison with one-step MPC.}. The gradient \(\nabla g(\mathbf{a}_t)\) captures how small changes in action would affect the goal, accounting for all encoded regularities simultaneously.

This is where the consequence of adaptive composition becomes most discernible. The gradient aggregates contributions from all active interconnections and recursive estimators, weighted by their current informativeness. When a regularity is highly informative, it contributes strongly to the gradient. When a regularity is uninformative or irrelevant to the current situation, its contribution diminishes. This yields context-sensitive behavior that emerges naturally from the network's adaptive composition, without requiring hand-designed switching rules, even for sequential and partially observable tasks~\cite{mengers_no_2025}.

\subsection{Summary and Preview}\label{sec:method-summary}

AICON realizes adaptive composition through three mechanisms: recursive estimators that enforce temporal consistency, active interconnections that enforce cross-variable relationships, and gradient descent that translates these regularities into action. The framework requires specifying \emph{which} regularities govern a problem, but not \emph{how} to combine them---that emerges automatically from the network's response to sensory feedback.

This raises an empirical question: does adaptive composition actually enable generalization? In the next section, we describe a controlled experimental setup designed to test this. We select a problem simple enough that all relevant regularities can be identified, construct an AICON model encoding those regularities, and then systematically expose it to conditions never considered during design.

\section{A Testbed for Adaptive Composition}\label{sec:setup}
To understand \emph{how} adaptive composition enables generalization, we require a problem in which all relevant regularities can be identified with confidence. This makes the mechanism visible: if the model generalizes, it must be because regularities combine appropriately, not because some unidentified regularity incidentally covers the new situation. We therefore select a deliberately simple problem: a robot maintaining a fixed distance from a target using only visual measurements. Our aim here is to understand, but not to benchmark. Prior work has demonstrated AICON's capability on physical robots and other domains~\cite{battaje_information_2024,mengers_no_2025,mengers_leveraging_2024,mengers_robotics-inspired_2025}.

\subsection{Base Problem and Governing Regularities}\label{sec:setup-base-problem}

We consider a simulated 2D environment in which a velocity-controlled holonomic robot equipped with a \ang{360} visual sensor must maintain a desired distance from a stationary target. Crucially, the robot has no direct measurement of distance. It must infer distance from visual information alone. The robot receives measurements of its own velocity in the local frame, \(\mathbf{z}^\mathrm{vel} = [v_x, v_y, \omega]^T\), and object-related visual measurements including the visual offset (bearing) \(\theta\), visual size \(\phi\), and their time derivatives: \(\mathbf{z}^\mathrm{target} = [\theta, \phi, \dot{\theta}, \dot{\phi}]^T\).

\begin{figure}[t]
	\includegraphics[width=\linewidth]{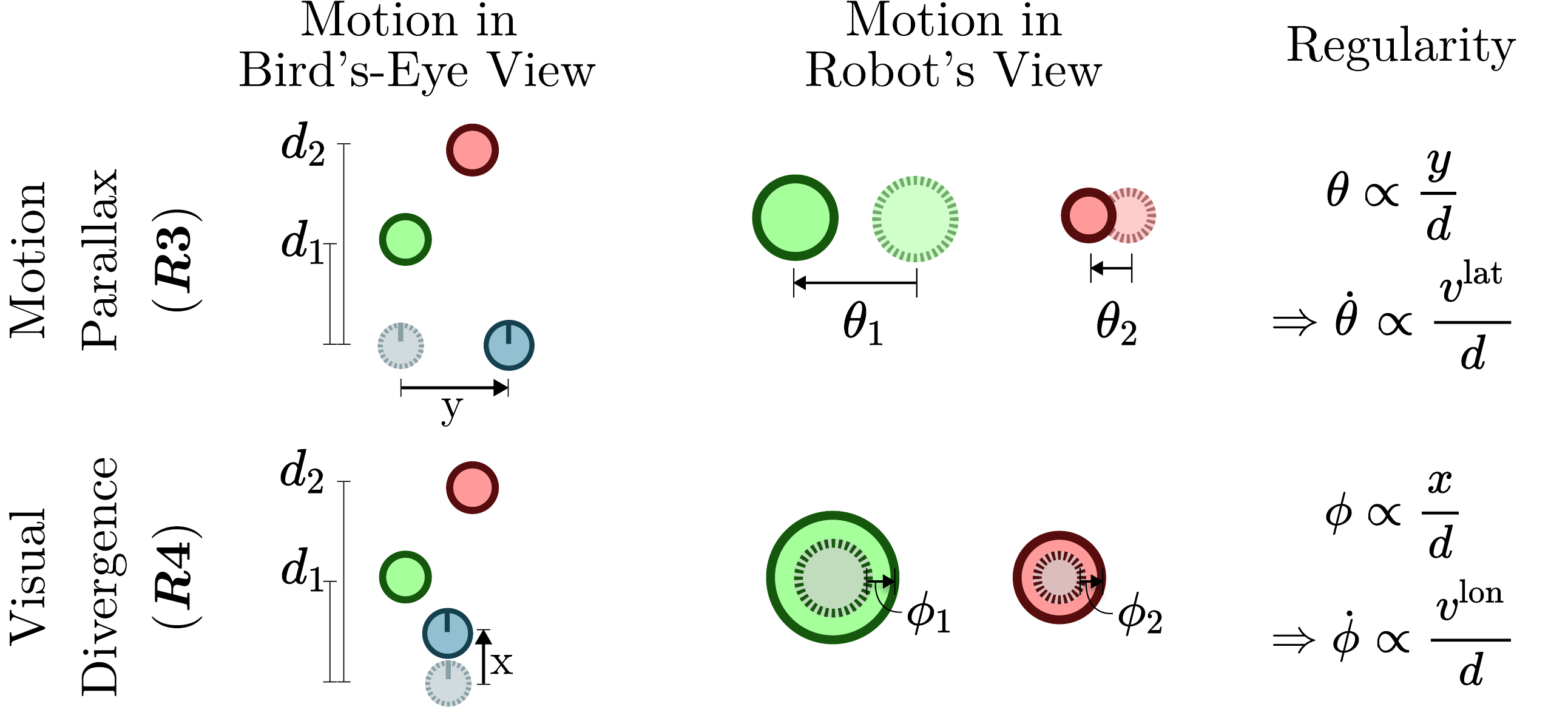}
	\caption{Two regularities enable distance estimation from visual measurements alone. \textbf{(Top)} When the robot moves laterally, nearby objects shift more in the visual field than distant ones---the rate of change in visual offset \(\dot{\theta}\) is inversely proportional to distance. \textbf{(Bottom)} When the robot moves longitudinally, nearby objects grow faster than distant ones---the rate of change in visual size \(\dot{\phi}\) is inversely proportional to distance. Left column shows bird's-eye view of robot motion; middle column shows resulting visual changes from the robot's perspective; right column shows the mathematical relationships (full relationships given in \Cref{eqn:R3-interconnection,eqn:R4-interconnection}).}
	\label{fig:explainer-smrs}
\end{figure}

This problem is governed by four regularities that together capture all relevant physical relationships:
{\setlength{\parindent}{\dimexpr 10pt-1ex\relax} 
	\paragraph*{\textbf{R1 -- Constant Velocity}} Over short intervals, both robot and object velocities can be assumed to be constant.
	\paragraph*{\textbf{R2 -- Reference Frame Consistency}} When the robot moves in some direction, a static object appears to move in the opposite direction by the same amount in the robot's reference frame.
	\paragraph*{\textbf{R3 -- Motion Parallax}} Lateral robot motions cause predictable changes in visual offset \(\dot{\theta}\) (See \Cref{fig:explainer-smrs}, top). Nearby objects shift more rapidly than distant ones.
	\paragraph*{\textbf{R4 -- Visual Divergence}} Longitudinal robot motions cause predictable changes in visual size \(\dot{\phi}\) (See \Cref{fig:explainer-smrs}, bottom). Approaching an object makes it appear larger at a rate inversely proportional to distance.
}

These four regularities are sufficient for the problem: \textit{\textbf{R1}}--\textit{\textbf{R2}} capture the rigid-body dynamics of robot and target, and \textit{\textbf{R3}}--\textit{\textbf{R4}} make distance observable from visual measurements given known robot velocity. Together they reveal the actual distance to the target, which can then be used to move to a specified distance---a strategy our hand-designed baseline controllers exploit explicitly~(\Cref{sec:setup-baselines}).

\subsection{Implementation in AICON}\label{sec:setup-aicon-model}

The regularities \textit{\textbf{R1}}--\textit{\textbf{R4}} map to AICON's structural components as follows. A schematic is provided in \Cref{fig:base-scenario}~(top-left).

\subsubsection{Recursive Estimators}
Two recursive estimators capture the temporal regularities. Based on \textit{\textbf{R1}}, one estimator tracks the robot's velocity \(\mathbf{x}^\mathrm{robot}_t = [v_{x,t}, v_{y,t}, \omega_t]^T\) given actions \(\mathbf{a}_t\):
\begin{equation}
    f^{\textit{\textbf{R1}}}(\mathbf{x}^\mathrm{robot}_t, \mathbf{a}_t) = \mathbf{x}^\mathrm{robot}_{t+1}.
\end{equation}

Based on \textit{\textbf{R1}} and \textit{\textbf{R2}}, the other estimator tracks \(\mathbf{x}^\mathrm{target}_t = [\theta^\mathrm{target}_t, d^\mathrm{target}_t, v^\mathrm{target}_{x,t}, v^\mathrm{target}_{y,t}, r^\mathrm{target}_t]^T\), representing the target's visual offset, distance, velocities, and size in the robot's frame:
\begin{equation}
    f^{\textit{\textbf{R2}}}(\mathbf{x}^\mathrm{target}_t; \mathbf{x}^\mathrm{robot}_t) = \mathbf{x}^\mathrm{target}_{t+1}.
\end{equation}

\subsubsection{Active Interconnections}

According to \textit{\textbf{R3}}, the robot velocity \(\mathbf{x}^\mathrm{robot}_t\), the target distance \(d^\mathrm{target}_t\) (from \(\mathbf{x}^\mathrm{target}_t\)), and the observed change in visual offset \(\dot{\theta}_t\) (from \(\mathbf{z}^\mathrm{target}_t\)) form a predictable relationship encoded in an active interconnection:
\begin{equation}
    c^{\text{\textit{\textbf{R3}}}}(\mathbf{x}^\mathrm{robot}_t, \mathbf{x}^\mathrm{target}_t, \mathbf{z}^\mathrm{target}_t) = \dot{\theta}_t + \frac{v^\mathrm{lat}_t}{d^\mathrm{target}_t} - \omega^\mathrm{robot}_t = 0,
	\label{eqn:R3-interconnection}
\end{equation}
where \(\omega^\mathrm{robot}_t\) is the robot's rotational velocity (from \(\mathbf{x}^\mathrm{robot}_t\)) and \(v^\mathrm{lat}_t\) is the lateral component of the relative velocity, obtained by rotating the robot and target velocity estimates by the estimated target visual offset \(\theta^\mathrm{target}_t\).

Following \textit{\textbf{R4}}, robot motion \(\mathbf{x}^\mathrm{robot}_t\) also relates to changes in visual size \(\dot{\phi}_t\) (from \(\mathbf{z}^\mathrm{target}_t\)) through the target's size and distance \(r^\mathrm{target}_t,\;d^\mathrm{target}_t\) (from \(\mathbf{x}^\mathrm{target}_t\)), encoded in another active interconnection:
\begin{equation}
    c^\text{\textit{\textbf{R4}}}(\mathbf{x}^\mathrm{robot}_t, \mathbf{x}^\mathrm{target}_t, \mathbf{z}^\mathrm{target}_t) = \dot{\phi}_t - \frac{2 \eta_t}{\sqrt{1 - \eta^2_t}} \frac{v^\mathrm{lon}_t}{d^\mathrm{target}_t} = 0,
	\label{eqn:R4-interconnection}
\end{equation}
where \(v^\mathrm{lon}_t\) is the longitudinal component of the relative velocity, obtained by rotating the robot and target velocity estimates by the estimated target visual offset \(\theta^\mathrm{target}_t\) (from \(\mathbf{x}^\mathrm{target}_t\)), and \(\eta_t = r^\mathrm{target}_t/d^\mathrm{target}_t\).

Direct measurements, such as robot velocity \(\mathbf{z}^\text{vel}_{t}\), visual offset \(\theta_t\) (from \(\mathbf{z}^\mathrm{target}_t\)) and size \(\phi_t\) (from \(\mathbf{z}^\mathrm{target}_t\)), are integrated into the respective estimators using simple one-to-one constraints.

\subsubsection{Goal Function and Behavior Generation}

To maintain a desired distance \(d^\mathrm{desired}\) from the target, we define a cost function on the target distance \(d^\mathrm{target}_t\) (from \(\mathbf{x}^\mathrm{target}_t\)):
\begin{equation}
    g(\mathbf{x}^\mathrm{target}_t) = \mathbb{E}\left[\left(d^\mathrm{target}_t - d^\mathrm{desired}\right)^2\right]. \label{eqn:goal-function}
\end{equation}

Behavior is generated by minimizing this cost using gradient descent with respect to motor commands \(\mathbf{a}_t\) through the adaptive composition, as described in \Cref{sec:method-gradients}.

\subsection{Baseline Controllers}\label{sec:setup-baselines}

We compare AICON against four baselines: two hand-designed policies that use the same regularities (isolating the contribution of adaptive composition) and two policies learned end-to-end with reinforcement learning (contrasting structured composition with direct policy learning):
{\setlength{\parindent}{\dimexpr 10pt-1ex\relax} 
	\paragraph*{\textbf{Parallax-Only}} Uses \textit{\textbf{R1}}-\textit{\textbf{R3}} to estimate distance through motion parallax. When target-distance uncertainty~(\(\sigma^\mathrm{target}\)) is high, lateral velocity is proportionally adjusted to reduce uncertainty. Longitudinal velocity is adjusted inversely proportional to uncertainty and proportional to distance error~(\(d^\mathrm{target} - d^\mathrm{desired}\)). This implements an explicit explore-exploit strategy.
	\paragraph*{\textbf{Fixed-Strategy}} Uses full estimation (\textit{\textbf{R1}}--\textit{\textbf{R4}}) with the same hand-crafted policy as above. This tests whether the regularities suffice for estimation and whether gradient-based action selection offers additional benefit.
	\paragraph*{\textbf{RL (PPO)}} A feed-forward policy trained with proximal policy optimization~\cite{schulman_proximal_2017} using dense rewards derived from the goal function~(\Cref{eqn:goal-function}). The model uses two-hidden-layer MLP with 16 units per layer ($\sim870$ trainable parameters total). Training uses only the stationary-target condition, but validation is done on both stationary and moving conditions. This tests whether learning a direct mapping from observations to actions matches structured composition.
	\paragraph*{\textbf{RL (RecurrentPPO)}} Same training setup with an LSTM-based recurrent policy (single LSTM layer of 8 hidden units feeding the same $16\times16$ MLP head; $\sim2{,}000$ trainable parameters total, $\sim2.3\times$ the feed-forward PPO), providing substantially more model capacity and temporal memory. This tests whether additional capacity and adding an inductive bias to model temporal evolution closes any remaining gap.
}

\subsection{Systematic Variations}\label{sec:setup-variations}

The base problem (\Cref{sec:setup-base-problem}) tests whether AICON can compose regularities appropriately for a single scenario. To test generalization more thoroughly, we introduce systematic variations that stress different aspects of the composition. Each variation modifies or extends the regularities \textit{\textbf{R1}}--\textit{\textbf{R4}}, requiring corresponding adjustments to the AICON model. Importantly, we do not redesign behavior for each variation---we only update the structural components to reflect the changed regularities and let behavior emerge through gradient descent.

{\setlength{\parindent}{\dimexpr 10pt-1ex\relax} 
	\paragraph*{\textbf{E1 -- Acceleration Control}} The robot is controlled via accelerations instead of velocities. The estimator for the robot's velocity is modified to account for accelerations 
	\(\mathbf{a_{t}} = [a_{x,t}^{action}, a_{y,t}^{action}, \dot{\omega}^{action}_t]^T\), which are integrated to update robot velocity under a constant-acceleration assumption. This tests whether composition handles altered actuation dynamics.
	
	\paragraph*{\textbf{E2 -- Differential-Drive Robot}} The robot uses differential-drive rather than holonomic motion. Actions \(\mathbf{a}_t = [v_{l,t}, v_{r,t}]^T\) correspond to left and right wheel velocities, mapped to overall motion through forward kinematics. This tests composition under constrained actuation.
	
	\paragraph*{\textbf{E3 -- Target Velocity Measurements}} The robot receives velocity measurements from the target---as in drone-to-person tracking with IMU-equipped targets---but loses measurements of its own velocity. This tests whether composition can exploit alternative information sources.
	
	\paragraph*{\textbf{E4 -- Restricted Field-of-View (FOV)}} The \ang{360} visual sensor is replaced with a limited FOV sensor. When the target falls outside the FOV, no measurements are received. Following~\cite[Sec.~V]{mengers_no_2025}, a scalar visibility likelihood \(p_\mathrm{visible}\) is estimated and incorporated into the active interconnections. When the target is not visible, the corresponding active interconnections are temporarily nonoperational. This tests composition under interrupted observability.
	
	\paragraph*{\textbf{E5 -- Camera-like Visual Measurements}} The \ang{360} visual sensor functions like LIDAR, but camera-based measurements couple the measurements of change in visual offset and visual size, even for \ang{180} fish-eye cameras stitched for \ang{360} view. We address this by combining the two active interconnections \(c^\text{\textit{\textbf{R3}}}\) and \(c^\text{\textit{\textbf{R4}}}\) into a single interconnection that preserves the original regularities while integrating the coupled visual measurements. This tests composition with altered sensor characteristics.
	
	\paragraph*{\textbf{E6 -- Obstacles in the Environment}} To incorporate possible obstacles, we add one recursive estimator and corresponding active interconnections per obstacle, analogous to the target estimator and its active interconnections. Measurements correspond to each obstacle's position and size. The goal function is modified to maintain a safe distance \(g(\mathbf{x}^\mathrm{\;obstacle}_{i,t}) = \mathbb{E}\left[\mathrm{min}(0,\;d^\mathrm{\;safety} - d^\mathrm{\;obstacle}_{i,t})^2\right]\),
	which enforces a safety margin \(d^\mathrm{\;safety}\) without affecting distant obstacles. This tests whether composition scales with more interacting constraints.
}

These variations span actuation (\textit{\textbf{E1}}, \textit{\textbf{E2}}), sensing (\textit{\textbf{E3}}, \textit{\textbf{E4}}, \textit{\textbf{E5}}), and environmental complexity (\textit{\textbf{E6}}). None were considered during original model design. Success across these variations would demonstrate that adaptive composition---rather than problem-specific engineering---drives generalization.

\begin{figure}[ht]
	\centering
	\includegraphics[width=\linewidth]{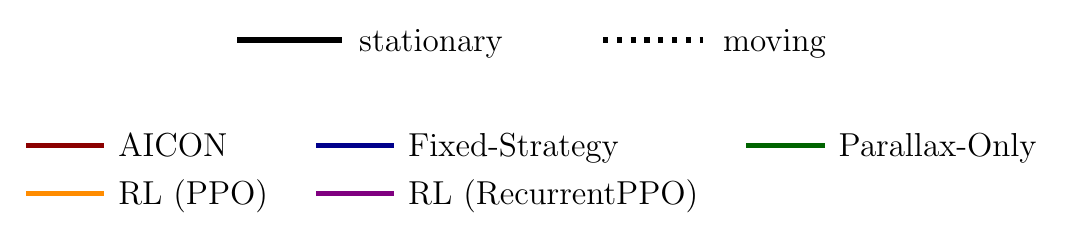}
	\includegraphics[width=\linewidth]{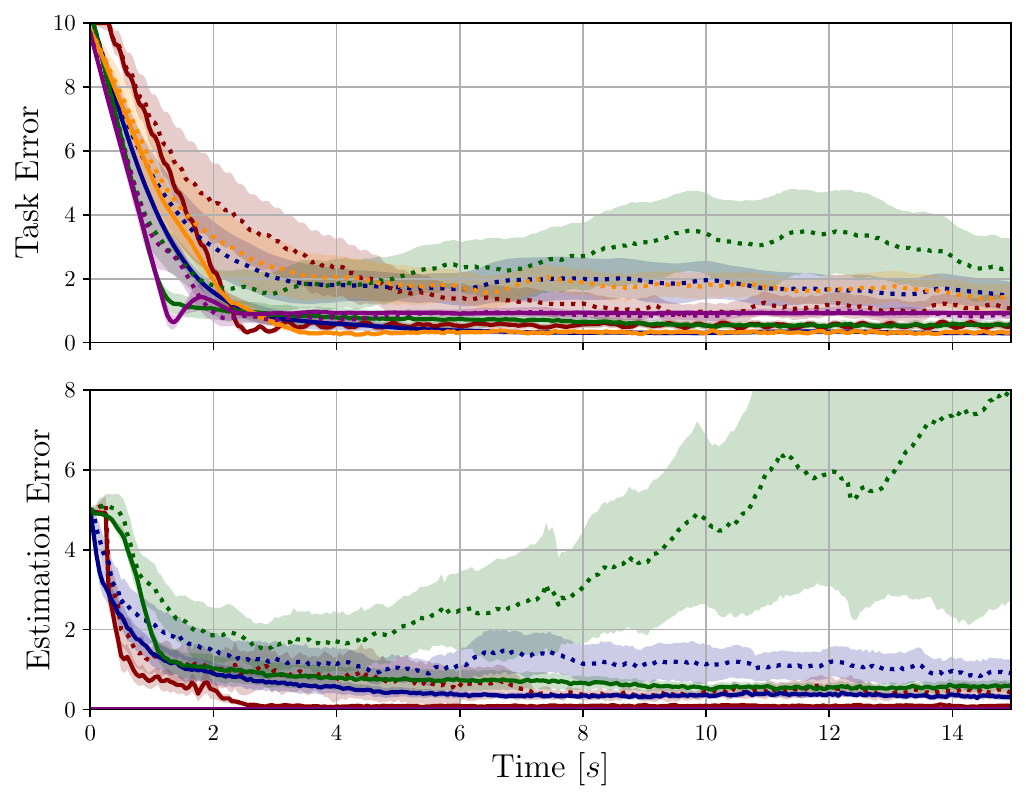}
	\caption{AICON generalizes to moving targets despite being designed only for stationary ones. \textbf{(Top)} Task error (distance from desired position) over time: For stationary targets (solid lines), all methods converge. For moving targets (dotted lines), AICON maintains low error through adaptive composition that continuously rebalances exploration and exploitation. RecurrentPPO matches AICON's performance asymptotically using substantially more model capacity, while PPO drops to Fixed-Strategy levels and Parallax-Only diverges. \textbf{(Bottom)} Estimation error over time, shown for estimator-based methods (RL baselines are end-to-end policies and have no explicit estimator): AICON and Fixed-Strategy maintain accurate estimates in both conditions, but Parallax-Only diverges for moving targets---its fixed explore-then-exploit strategy cannot adapt to the changing situation. Shaded regions indicate 99\% confidence intervals across 100 trials.} 
	\label{fig:baseline-comparison}
\end{figure}

\section{Empirical Results}\label{sec:results}

We now examine whether adaptive composition of regularities enables generalization. Our experiments demonstrate three findings of increasing scope: (A) composition enables the task and generalizes beyond design conditions; (B) composition adapts automatically, modulating regularity influence based on informativeness; and (C) generalization persists even when regularities must change, failing only when they become fundamentally insufficient.

\subsection{Composition Enables the Task}\label{sec:results-baseline}

\begin{figure*}[ht]
	\centering
	\includegraphics[width=0.98\textwidth]{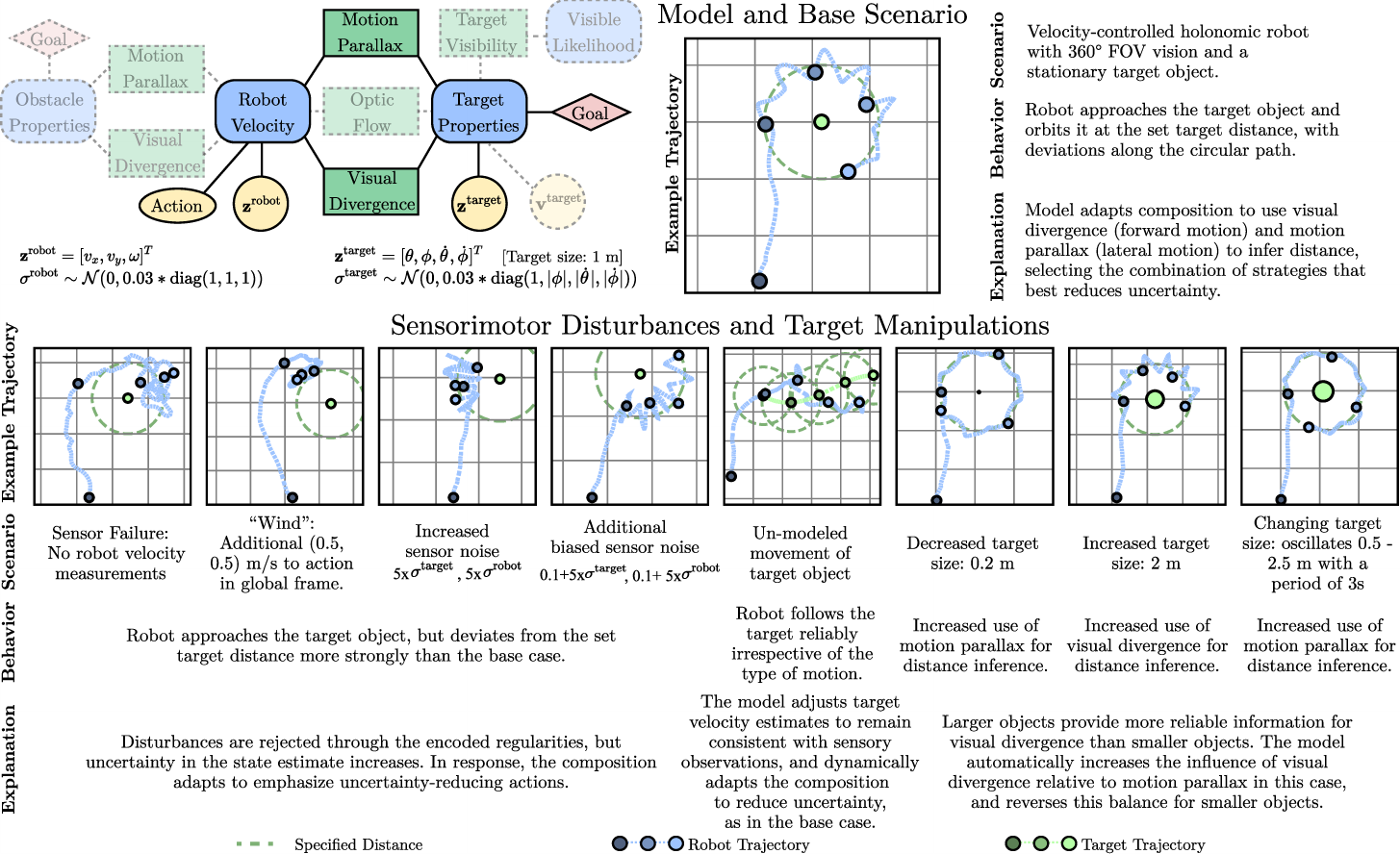}
	\caption{Adaptive composition generates appropriate behavior across diverse conditions without model modification. \textbf{(Top left)} The AICON model composes regularities \textit{\textbf{R1}}--\textit{\textbf{R4}} through two recursive estimators (Robot Velocity, Target Properties) linked by active interconnections (Motion Parallax, Visual Divergence). Grayed-out components indicate extensions described in \Cref{sec:setup-variations}, and are not useful to interpret the results in this figure. \textbf{(Top right)} In the base scenario, the robot approaches and maintains the desired distance from a stationary target, balancing exploration (lateral motion for distance estimation) with goal pursuit (longitudinal motion). \textbf{(Bottom)} Under sensorimotor disturbances (left four panels) and target manipulations (right four panels), the same unmodified model generates situation-appropriate behavior. Disturbances increase estimation uncertainty, producing stronger corrective actions; target size variations automatically shift the balance between motion parallax and visual divergence strategies (see \Cref{sec:results-ablation}).}
	\label{fig:base-scenario}
\end{figure*}

We first test the AICON model against baselines under two conditions: stationary targets (matching design assumptions) and moving targets (violating them). For each of ten test cases (five models \(\times\) two conditions), we run 100 experiments with randomized initial positions, each lasting 300 timesteps, sufficient for baseline controllers to converge.

The moving target condition reveals the key finding: nothing in the AICON model explicitly encodes for a moving target, yet AICON maintains low task error (\Cref{fig:baseline-comparison}, top, dotted lines). The hand-crafted baselines fail to adapt---Parallax-Only diverges and Fixed-Strategy plateaus at higher error. The learned baselines reveal a complementary picture: PPO trained only on stationary targets drops to Fixed-Strategy levels when the target moves, while RecurrentPPO, with substantially more model capacity, asymptotically matches AICON. AICON's gradient-based action selection that uses only four regularities automatically rebalances exploration and exploitation based on current estimation quality (\Cref{fig:baseline-comparison}, bottom): zero-shot generalization arises because its structure mirrors the regularities governing the problem, rather than from training on hundreds of episodes using far greater number of parameters.

\Cref{fig:base-scenario} illustrates a sample of the generated behavior and shows that the same unmodified model handles sensorimotor disturbances and target manipulations by generating movement patterns suited to each condition.

These results establish that adaptive composition not only enables the task but provides both robustness to sensorimotor disturbances and generalization to novel target manipulations---capabilities that extend beyond the initial design conditions.

\subsection{Composition Adapts Automatically}\label{sec:results-ablation}

A stronger test asks whether the network automatically recalibrates when its structure changes. We investigate this by ablating the active interconnections encoding \textit{\textbf{R3}} and \textit{\textbf{R4}}.

When we remove the motion parallax interconnection (\textit{\textbf{R3}}), the network can no longer reduce distance uncertainty through lateral motion. The model automatically compensates by producing longitudinal motion that exploits visual divergence (\textit{\textbf{R4}}) instead. The converse occurs when we ablate visual divergence: the model shifts to lateral motion exploiting motion parallax. Neither compensatory strategy was designed---both emerge from gradient descent through the modified network.

\begin{figure*}[ht]
	\includegraphics[width=\linewidth]{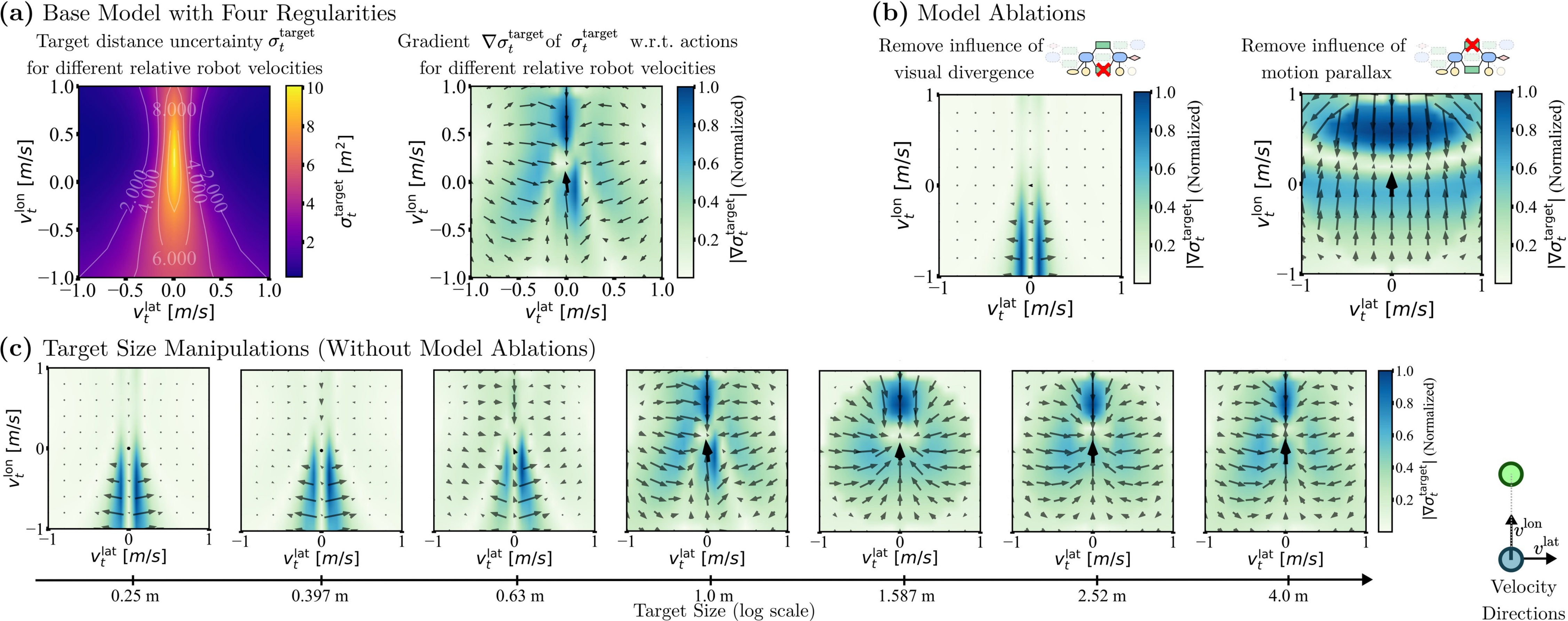}
	\caption{Adaptive composition automatically modulates which regularities influence behavior based on their current informativeness. 
	\textbf{(a)} Left plot shows how different robot action velocities (relative to target) lead to different distance uncertainties. Right plot shows the gradient of uncertainty, with arrows indicating directions of increasing uncertainty. This gradient directly specifies what action minimizes uncertainty for any configuration.
	\textbf{(b)} Ablating visual divergence (\textit{\textbf{R4}}) or motion parallax (\textit{\textbf{R3}}) shifts behavioral preferences toward lateral or longitudinal motion, as the model exploits ``remaining'' regularities. Model configurations are shown for each condition. Find full model schematic in \Cref{fig:base-scenario}~(top).
	\textbf{(c)} Target size variations produce similar patterns on a continuum: small targets favor lateral motion (motion parallax more informative) while large targets favor longitudinal approaches (visual divergence more informative).
	All analyses use identical robot-target configurations except for action and target size (panel c).
	}
	\label{fig:ablation}
\end{figure*}

\Cref{fig:ablation} visualizes this adaptation by plotting the gradient of distance uncertainty with respect to actions\footnote{The goal function~(\Cref{eqn:goal-function}) decomposes into two terms: \(g(\mathbf{x}^\mathrm{target}_t) = \left(d^\mathrm{target}_t - d^\mathrm{desired}\right)^2 + \sigma^\mathrm{target}_t\). A least-squares distance term---for which the optimal action is simply to move toward or away from the target---and an uncertainty term \(\sigma^\mathrm{target}_t\) where both \textit{\textbf{R3}} and \textit{\textbf{R4}} contribute. We visualize only the latter to highlight the more interesting behavioral consequences.}. The full model (\Cref{fig:ablation}a) shows how different actions lead to different uncertainty outcomes. When we ablate regularities (\Cref{fig:ablation}b), the gradient landscape shifts: removing \textit{\textbf{R4}} creates a preference for lateral motion, while removing \textit{\textbf{R3}} favors longitudinal motion.

A surprising finding emerged when we varied target size without ablation (\Cref{fig:ablation}c). Small targets produce gradient landscapes resembling the \textit{\textbf{R4}}-ablated case, while large targets resemble the \textit{\textbf{R3}}-ablated case. Intermediate sizes show continuous transitions. This makes physical sense: approaching a large object produces greater change in visual angle than approaching a small one, making visual divergence more informative for large targets and motion parallax more informative for small ones. We did not anticipate this behavior during design---it emerges entirely from the interactions between regularities.

These results demonstrate that adaptive composition is not merely a static encoding of regularities but a dynamic process that automatically modulates the influence of each regularity based on its current utility.

\subsection{Generalization Persists Under Modified Regularities}\label{sec:results-extensions}

\begin{figure*}[ht]
	\centering
	\includegraphics[width=0.98\textwidth]{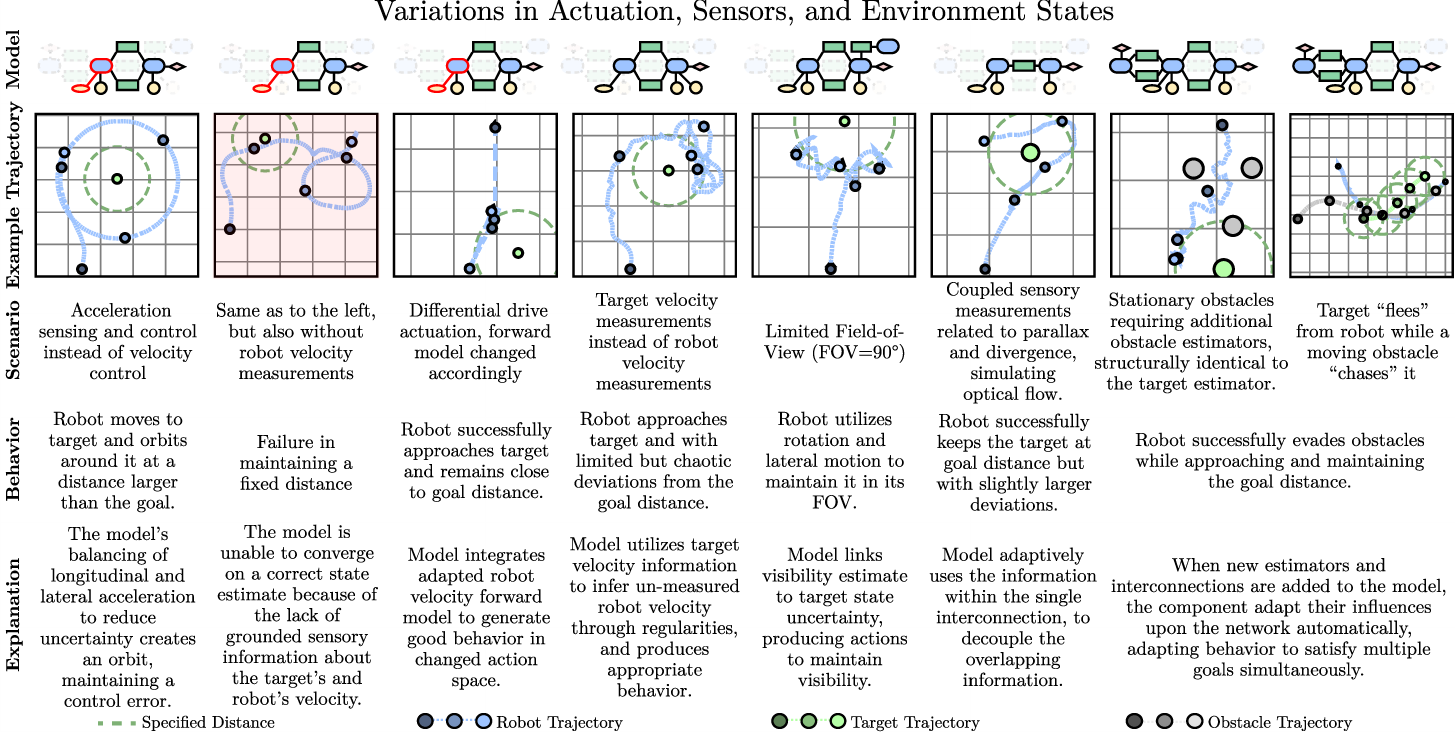}
	\caption{Generalization persists when regularities are modified, and fails precisely when they become insufficient. Each column shows a variation (\textit{\textbf{E1}}--\textit{\textbf{E6}} and combinations) requiring changes to the model structure (top row, model components as shown in \Cref{fig:base-scenario} but modified (red outline) or dis-/enabled (opacity)). Despite these changes to the network, appropriate behavior emerges: the differential-drive robot compensates for lost lateral motion through rotation; the restricted-FOV robot actively maintains target visibility; the obstacle scenario balances approach and avoidance. The single failure (second column, red background) occurs when acceleration control is combined with removal of velocity measurements---here, the regularities are fundamentally insufficient because integrating acceleration cannot determine absolute velocity without a reference. This failure confirms our proposition: composition enables generalization exactly to the extent that the encoded regularities capture the relevant structure.}
	\label{fig:extensions}
\end{figure*}

The strongest test of generalization is when some of the regularities themselves are modified while others remain static. We carry this out using six extensions (\textit{\textbf{E1}}--\textit{\textbf{E6}}, \Cref{sec:setup-variations}).

\Cref{fig:extensions} presents the results across all extensions. We observe that the generated behaviors are remarkably well-suited to the conditions. For instance, when the robot actuation is changed to differential-drive, it compensates for the lack of lateral movement by rotating the target to one side, moving forward, and then reorienting toward the target as needed. Combinations of extensions (e.g., differential-drive with obstacles, restricted FOV with camera-like measurements) also produce appropriate behavior without any additional engineering.

\subsubsection*{When Composition Fails}
One condition produced failure: acceleration control (\textit{\textbf{E1}}) \emph{without} robot velocity measurements. Integrating acceleration alone cannot determine absolute velocity, as the integration constant remains unspecified. Consequently, estimation becomes ungrounded, and behavior fails.

This failure is instructive: it occurs not because of a limitation in the composition mechanism, but because the encoded regularities are fundamentally insufficient. No amount of adaptive recombination can recover unavailable information. Generalization succeeds when regularities capture the relevant structure, and fails when they do not.

\subsection{Results Summary}

Across 17 tested scenarios, AICON generates context-appropriate behavior in all but one case. This success stems from how adaptive composition exploits whatever regularities remain informative in each situation. The ablation and target-size experiments reveal the mechanism: both manipulations produce identical shifts in behavioral strategy because both alter which regularities carry useful information. The single failure---acceleration control without velocity measurements---confirms the boundary: adaptive composition cannot compensate when regularities lack the information needed for the task.

\section{Discussion \& Conclusion}\label{sec:discussion-and-conclusion}

Our empirical study set out to test a specific proposition: that adaptive composition of regularities enables generalization. The results show that it works across a wide range of conditions. More importantly, our experiments reveal \emph{why} this works.

Regularities reliably describe the structure of robot-environment relationships. Adaptive composition determines how they combine in each situation. This means that the designer specifies the regularities, and how they combine emerges from the network's response to sensory feedback. The target size experiment illustrates this clearly. We encoded motion parallax and visual divergence without knowing that target size would affect their relative utility. Yet the composition leveraged this structure automatically. The designer's task thus shifts from crafting solutions to characterizing structure.

\subsubsection*{Toward Compositional Learning}
Most regularities governing robot-environment interactions are physical relationships we already understand. But understanding does not always yield closed-form expressions. Consider non-rigid body dynamics: we understand the underlying physics well enough to simulate fabric, yet characterizing its behavior in compact form remains elusive. Such regularities could be learned from data, and nothing in our results precludes this. Learning regularities individually and embedding them within a compositional architecture could enable efficient acquisition of complex behaviors. This would proceed \emph{regularity-by-regularity} rather than \emph{end-to-end}, retaining the generalization benefits observed here while reducing manual engineering.

Identifying which regularities to encode is itself an open problem, distinct from expressing them once chosen or learning them. Ongoing work points in two directions. Recursive estimators with outlier-detection mechanisms can reveal when the current composition is insufficient. A recurring outlier characterizes the conditions under which the composition does not capture the world structure. This can in turn indicate where a new active interconnection or estimator should be placed, through a mix of manual analysis and automated discovery. Separately, an AICON network can act as a structured prior over task dynamics for reinforcement and supervised learning: pairing it with a compact learned policy that modulates its gradient paths lets task learning inherit the structural generalization of the composition instead of acquiring it from scratch~\cite{sebastian_world-task_2026}.

\subsubsection*{Toward Generalization in Robotics}
Generalization remains the central challenge in robotics. Our results suggest that adaptive composition of regularities offers a promising path forward and constitutes a valuable inductive bias for behavior generation. This insight need not be tied to AICON specifically. What matters is that the world's structure arises from regularities that combine differently across situations. Systems that mirror this structure can therefore generalize in a principled way.

In this paper, we studied a single simulated domain in depth so the mechanism of adaptive composition could be examined under controlled conditions. Increased confidence in the versatility of adaptive composition will require testing across other domains. Prior work has applied AICON to physical robots~\cite{mengers_no_2025}, collective estimation~\cite{mengers_leveraging_2024}, and modeling in psychology~\cite{battaje_information_2024,mengers_robotics-inspired_2025}, though without explicitly testing for generalization. Given those results alongside ours, we expect adaptive composition to yield similar benefits in other behavior generation problems. More broadly, this perspective could reshape how robotics problems are approached: any problem can be framed by asking what regularities govern it and how they should compose. Advances in generalization may then come not from larger models or more data, but from better understanding of the structure we aim to capture. Such understanding offers a foundation for building robotic systems that generalize not by memorizing situations, but by understanding the regularities that generate~them.

\section*{Acknowledgments}
We gratefully acknowledge funding by the Deutsche Forschungsgemeinschaft (DFG, German Research Foundation) under Germany's Excellence Strategy -- EXC 2002/1 ``Science of Intelligence'' -- project number 390523135. This work has been partially supported by the German Federal Ministry of Research, Technology and Space (BMFTR) under the Robotics Institute Germany~(RIG).

\newpage

\bibliographystyle{IEEEtran-nourl}
\bibliography{Battaje-Bernhard-Mengers-26-ICRA}

\end{document}